\documentclass[10pt,twocolumn,letterpaper]{article}

\usepackage{btas}
\usepackage{times}
\usepackage{epsfig}
\usepackage{graphicx}
\usepackage{amsmath}
\usepackage{amssymb}
\usepackage{subfigure}
\usepackage{float}
\usepackage{multirow}
\usepackage{multicol} 
\usepackage{xcolor}
\usepackage{colortbl}  %
\usepackage{url}
\usepackage{eurosym}
\usepackage{setspace}  
\usepackage{caption}
\usepackage{pifont}
\usepackage{threeparttable} 
\usepackage{enumerate}
\usepackage{url}

\btasfinalcopy 

\ifbtasfinal\fi
\begin{document}

\title{A database for face presentation attack using wax figure faces}

\author{Shan Jia, Chuanbo Hu, Guodong Guo, and Zhengquan Xu\\
Wuhan Uiniversity, China, and West Virginia University, USA\\
{\tt\small jias@whu.edu.cn, Guodong.Guo@mail.wvu.edu}
}

\maketitle
\thispagestyle{empty}

\begin{abstract}
   Compared to 2D face presentation attacks (e.g. printed photos and video replays), 3D type attacks are more challenging to face recognition systems (FRS) by presenting 3D characteristics or materials similar to real faces. Existing 3D face spoofing databases, however, mostly based on 3D masks, are restricted to small data size or poor authenticity due to the production difficulty and high cost. In this work, we introduce the first wax figure face database, WFFD, as one type of super-realistic 3D presentation attacks to spoof the FRS. This database consists of 2200 images with both real and wax figure faces (totally 4400 faces) with a high diversity from online collections. Experiments on this database first investigate the vulnerability of three popular FRS to this kind of new attack. Further, we evaluate the performance of several face presentation attack detection methods to show the attack abilities of this super-realistic face spoofing database.
\end{abstract}

\section{Introduction}
With the widespread face recognition technologies, the security and privacy risks of face recognition systems (FRS) have been increasingly become as a critical issue in both academia and industry. Face presentation attacks are one of the most easily realized threats by presenting an artificial object or a copy or synthetic pattern of faces to the biometric data capture subsystem~\cite{ISO2016}. Based on the way to generate the face artifact, face presentation attacks can be classified into 2D modalities (which present printed/digital photographs or recorded videos on the mobile/tablet), and 3D type (by wearing a mask or presenting a synthetic model).

Because of the simplicity, efficiency, and low cost of making 2D type attacks, current systems and research pay more attentions to 2D face presentation attacks. However, with similiar 3D structures or materials to real faces, 3D face presentation attacks are more hyper-realistic, and therefore, more powerful to attack the FRS while more difficult to detect. For example, as shown in Figure 1, even some systems, which have already taken presentation attack detection (PAD) into consideration, can be fooled by the 3D face presentation attacks. 

\begin{figure}[t]
\begin{center}
\subfigure[]{\label{(a)}
\includegraphics[width=1.35in,height=1.in]{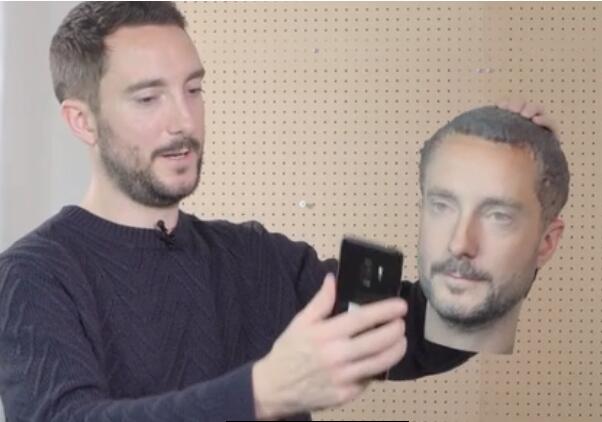}}
\subfigure[]{\label{(b)}
  \includegraphics[width=1.35in,height=1.in]{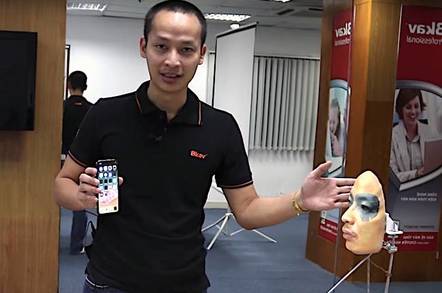}}
\caption{Examples of 3D presentation attack cases. (a) Android phones fooled by a 3D-printed head\protect\footnotemark[1], (b) iPhone X face ID unlocked by a 3D mask\protect\footnotemark[2].}
\label{fig:01}  
\end{center}
\end{figure}
\footnotetext[1]{Picture is downloaded from \url{https://www.forbes.com/sites/thomasbrewster/2018/12/13/we-broke-into-a-bunch-of-android-phones-with-a-3d-printed-head/#1de03dee1330.}}
\footnotetext[2]{Picture is downloaded from \url{https://www.theregister.co.uk/2017/11/28/iphone_x_face_id_system_cracked_again/.}}

Existing 3D presentation attacks are mostly based on wearable face masks. 3D facial mask spoofing was previously thought impossible to become a common practice in the literature~\cite{zhang2012face}, because compared to 2D type attacks, 3D masks are much more difficult and high-cost to manufacture, requiring special 3D devices and materials. In recent years, the rapid advancement of 3D printing technologies and services has made it easier and cheaper to make 3D masks. Several 3D mask attack databases have been created, inlcuding the 3DMAD (3D Mask Attack Database~\cite{3DMAD2013spoofing}), the 3DFS-DB (3D-face spoofing database~\cite{galbally2016three}), the HKBU-MARs (HKBU 3D Mask Attack with Real World Variations Database~\cite{liu20163d}), and the SMAD (Silicone Mask Attack Database~\cite{manja2017detecting}).

Although these 3D presentation attack databases have collected different 3D masks based on the third-party services~\cite{3DMAD2013spoofing, liu20163d}, by self-manufacturing~\cite{galbally2016three}, or from online resources~\cite{manja2017detecting}, the databases are restricted to small data sizes (mostly less than 30 subjects) or low mask qualities. This will not only limit the attack abilities of these fake faces, but also limit research works in reporting robust detection performance against 3D presentation attacks. 

To address this problem, we take advantage of the popularity and publicity of numerous celebrity wax figure museums in the world, and collect a large number of wax figure images to form the new Wax Figure Face Database (WFFD). These life-size wax figure faces are all carefully designed and made in clay with wax layers, silicone or resin materials, so that they are super-realistic and similar to real faces. With the development of wax figure manufacture technologies and services, we think the easily obtainable and super-realistic wax figure faces will pose threat to the face recognition systems. Therefore, we introduce these wax figure faces as a new challenging type of 3D face presentation attacks in this paper, and analyze their impact on face recognition. 

To the best of our knowledge, this is the first wax figure face database, also with a large data size and high diversity. Altogether, the WFFD consists of 2200 images of 450 subjects (with both real and wax figure faces, totally 4400 faces), which are diversified in subject age, ethnicity, face pose, facial expression, recording environment, and cameras, and therefore closer to the practical situation. Based on the new database, we first investigate the vulnerability of three popular face recognition systems to these super-realistic presentation attacks, and further evaluate the performance of several popular face PAD methods to show the attack ability of this 3D face spoofing database.

The rest of this paper is organized as follows. In Section 2, we introduce the related research in 3D face presentation attack databases and PAD methods. The new WFFD database is presented in Section 3. Section 4 describes the experiments and results on the proposed database. Finally, Section 5 concludes the paper.

\section{Related work}

\subsection{Existing 3D face presentation attack databases}

Most existing 3D face presentation attack databases create attacks by presenting wearable face masks, which are made of different materials with similar 3D face characteristics to the real faces. 3DMAD~\cite{3DMAD2013spoofing} is the first publicly available 3D mask database. It used the services of ThatsMyFace\footnote[3]{http://thatsmyface.com/.} to manufacture 17 masks of users, and recorded 255 video sequences with an RGB-D camera of Micsoft Kinect device for both real access and presentation attacks. This database is widely used by providing color images, depth images, and manually annotated eye positions of all face samples. 

With the development of 3D modeling and printing technologies, from 2016, more mask databases were created. 3DFS-DB~\cite{galbally2016three} is a self-manufactured and gender-balanced 3D face spoofing database. They made 26 printed models using two 3D printers: the ShareBot Pro and the CubeX\footnote[4]{https://www.sharebot.it. and  http://www.cubify.com.}, which are relatively low-cost and worth about 1000 and 2000 \euro, respectively. Acrylonitrile Butadiene Styrene (ABS) plastic material is used to generate the physical artifacts. HKBU-MARs~\cite{liu20163d} is another 3D mask spoofing database with more variations to simulate the real world scenarios. It generated 12 masks from two companies (ThatsMyFace and REAL-F\footnote[5]{http://real-f.jp/en\_the-realface.html.}) with different appearance qualities. 7 camera types and 6 typical lighting settings are also included to form totally 1008 videos. To include more subjects, the SMAD database~\cite{manja2017detecting} collected and compiled videos of people wearing silicone masks from online resources. It contains 65 genuine access videos of people auditioning,
interviewing, or hosting shows, and 65 attacked videos of people wearing a complete 3D structure (but not customized) mask around the head which fits well with proper holes for the eyes and mouth. 

Besides, there have been some 3D mask spoofing databases with special lighting information for more effective detection. The MLFP database~\cite{agarwal2017face} (Multispectral Latex Mask based Video Face Presentation Attack database) is a unique multispectral database for face presentation attacks using latex and paper masks. It contains 1350 videos of 10 subjects in visible, near infrared (NIR), and thermal spectrums, which are captured at different locations (indoor and outdoor) in an unconstrained environment. Similarly, the ERPA database~\cite{bhat2017you} also provides the RGB and NIR images of both bona fide and 3D mask attack presentations captured using special cameras. This is a small dataset with frame images of 5 subjects stored. The depth information is also provided. Both rigid resin-coated masks and flexible silicone masks are considered. 

These databases have played a significant role in designing multiple detection schemes against 3D face presentation attacks. However, they still face the problems of small database size, low diversity, or poor authenticity, which will certainly limit the development of effective and practical detection schemes. 

\subsection{3D face PAD methods}
Due to the smaller texture defects, better preserved motion, and more geometric properties in 3D face presentation attacks, detection of 3D fake faces is more challenging than detecting fake faces with 2D planar surfaces. Existing PAD methods for 3D face presentation attacks can be broadly classified into five categories, namely, the reflectance properties based, texture based, shape based, deep features based, and liveness based methods.

The earliest studies in 3D mask spoofing detection were based on the reflectance difference of facial skins and mask materials~\cite{kim2009masked, zhang2011face, wang2013new, steiner2016reliable}. This kind of methods can achieve a good detection performance, but one main limitation is the requirements of special and expensive devices to acquire multispectral images at various wavelengths. Texture based methods, differently, explore the texture pattern difference of real faces and masks with the help of texture feature descriptors, including the widely used Local Binary Patterns (LBP)~\cite{MORPHO1kose2013, MORPHO2kose2013shape,erdog2014spoofing, 3DMAD2013spoofing}, the Binarized Statistical Image Features (BSIF)~\cite{raghavendra2014novel, naveen2016face}, and Haralick features~\cite{agarwal2016face}. These methods are easy to implement, but their robustness to different mask spoofing attacks needs further investigations. Shape based 3D mask PAD methods use shape descriptors~\cite{MORPHO3kose2013vulnerability, tang20173d, hamdan2017detection} or 3D reconstruction~\cite{wang2018face} to extract discriminative features from faces and 3D masks. Instead of extracting hand-crafted features, deep feature based methods~\cite{menotti2015deep, lucena2017transfer, shao2017deep, liu2018detecting} automatically extracts features from the face images, and trend to have a higher detection accuracy and a better generalization ability.

Recently, some methods explore liveness cues to detect 3D face presentation attacks, such as thermal signatures~\cite{bhat2017you}, gaze information~\cite{ali2017biometric, alsufyani2018biometrie, ali2018gaze}, and pulse or heartbeat signals~\cite{liu20163d2, liu2018remote, hern2018time, Li2017Generalized}. Based on the intrinsic liveness signals, these methods achieve an outstanding performance in distinguishing real faces from masks.


\section{The wax figure face database}
To address the issues in existing 3D face presentation attack databases, we introduce the new WFFD database with a large size and high diversity as super-realistic 3D presentation attacks. The details of the data collection process, and the evaluation protocols are presented in this section.

This Wax Figure Face Database is based on numerous celebrity wax figure images from online resources. These user-customized and life-size wax figure faces are all carefully designed and made in clay with wax layers, silicone or resin materials, so that they are super-realistic. We first downloaded multiple celebrity wax figure faces as attacks with a high diversity in subject age, ethnicity, face pose, expression, recording environment, and cameras, and then collected the corresponding celebrity images as real access attempts. For each subject, the wax figure face and real face were finally grouped in one image to show the high authenticity, as the examples shown in Figure 2. 
\begin{figure}[h]
\centering
\includegraphics[height=3.9cm]{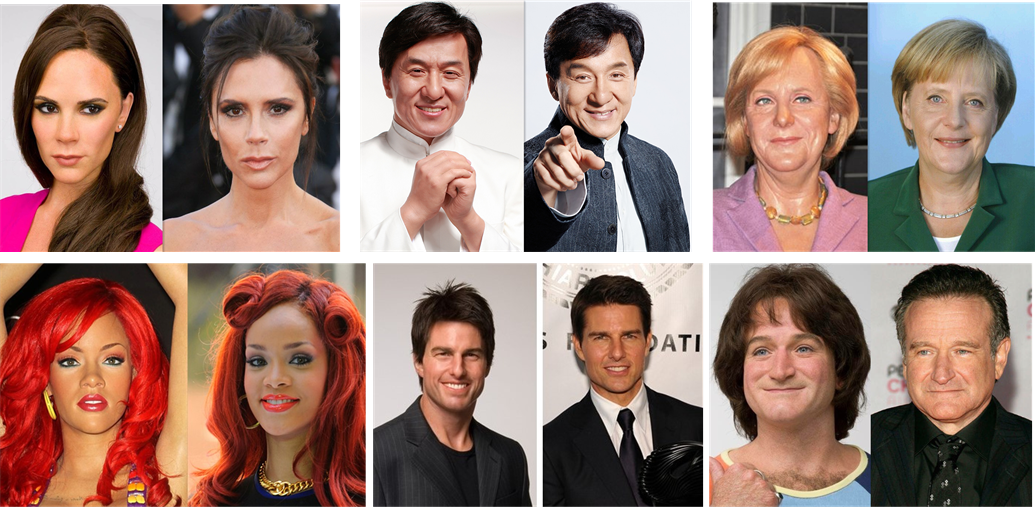}
\caption{Examples of images in the Protocol I of WFFD database.}
\end{figure}

Further, we introduce one more challenging scenario where the wax figure face and real face were originally recorded together, as shown in Figure 3. With the same recording environment, and even the same face poses and facial expressions, these images are more difficult to distinguish. 
\begin{figure}[h]
\centering
\includegraphics[height=5.4cm]{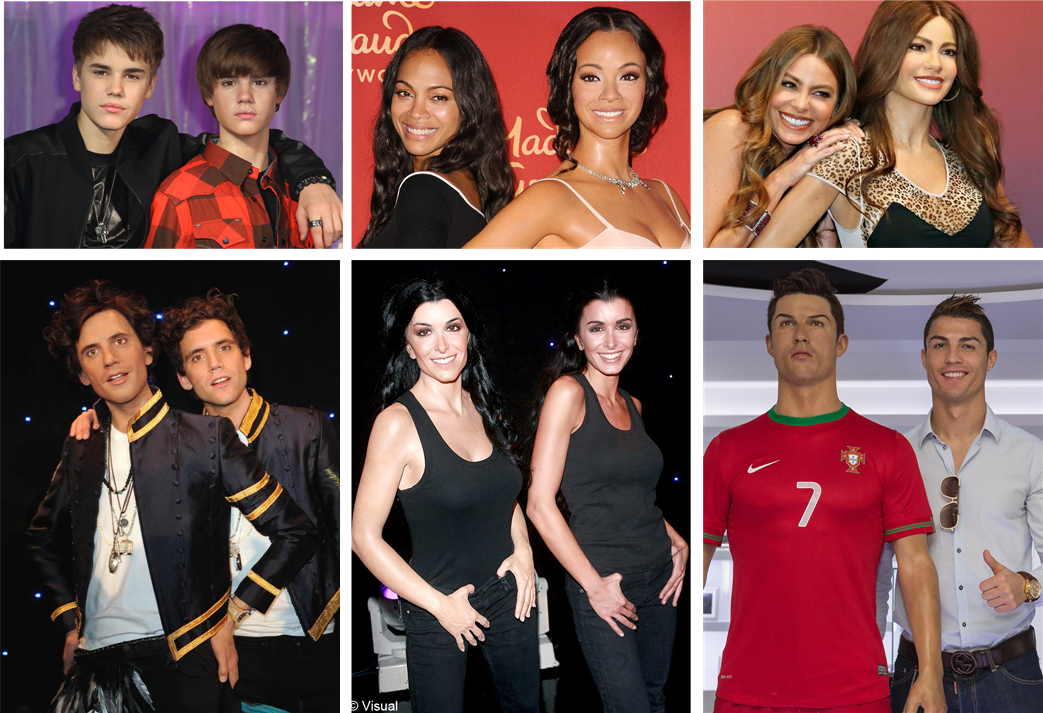}
\caption{Examples of images in the Protocol II of WFFD database.}
\end{figure}

Altogether, the WFFD consists of 2200 images with both real and wax figure faces of 470 subjects, totally 4400 faces. Table 1 compares the characteristics of WFFD with six existing 3D face presentation attack databases. To evaluate the performances of the face PAD methods on the WFFD database, we further designed three protocols.

1) Protocol I : This protocol contains images grouped manually, which means the wax figure face and real face came from different recording devices and environment.

2) Protocol II: Images in this protocol record the wax figure face and real face in the same environment with the same cameras.

3) Protocol III: the previous two protocols are combined to simulate the real-world operational conditions.

\begin{table*}[h]
\newcommand{\tabincell}[2]{\begin{tabular}{@{}#1@{}}#2\end{tabular}} 
\renewcommand{\arraystretch}{1.3}
\setlength{\tabcolsep}{5pt}
\begin{threeparttable}   
\caption{Comparison of 3D face presentation attack databases}
\centering
\begin{tabular}{|l|c|c|c|c|c|c|}
  \hline 
  \textbf{Database}&\textbf{Year}&\textbf{\#Subject}&\textbf{\#Sample}&\textbf{Format}&\textbf{Material}&\textbf{Description}\\   
  \hline 
  3DMAD~\cite{3DMAD2013spoofing} &2013&17&255 &video &paper, hard resin & 2D color images + 2.5D depth maps\\
   \hline
  3DFS-DB~\cite{galbally2016three}&2016&26 &520 &video &plastic & 2D, 2.5D images + 3D information\\ 
   \hline
  HKBU-MARs~\cite{liu20163d}&2016&12 &1008&video &/ &color images\\  
   \hline  
  SMAD~\cite{manja2017detecting} &2017&/&130& video &silicone & color images, from online resources\\    
   \hline
  MLFP~\cite{agarwal2017face} &2017&10&1350 &video &latex, paper & visible, NIR, thermal images\\ 
   \hline
  ERPA~\cite{bhat2017you} &2017&5&86 &image &resin, silicone & RGB, thermal, NIR images + depth\\  
   \hline
  WFFD (proposed)&2019 &470 & 2200 &image & wax figure & high diversity, from online resources\\
  \hline
\end{tabular}
\end{threeparttable}
\end{table*}

More details about the images used in each protocol are shown in Table 2.
\begin{table}[h]
\renewcommand{\arraystretch}{1.10}
\setlength{\tabcolsep}{3pt}
\caption{Number of images in each evaluation protocol}
\centering
\begin{threeparttable}   
\begin{tabular}{|l|c|c|c|c|}
  \hline
  \textbf{Protocol}&\textbf{\#Train}&\textbf{\#Development}&\textbf{\#Test}&\textbf{\#Total}\\  
  \hline
  Protocol I &600 &200 &200 &1000 \\   
  \hline
  Protocol II &720 &240 & 240 &1200 \\
  \hline
  Protocol III &1320&440 &440 &2200\\
  \hline
\end{tabular}
\begin{tablenotes}
\scriptsize
\item[*] Note that the train, development, and test subsets have no overlap.
\end{tablenotes}
\end{threeparttable}
\end{table}

\section{Experiments}
In this section, we first investigate the vulnerability of three popular face recognition systems to these super-realistic 3D presentation attacks, and then evaluate the performance of several popular face PAD methods on this 3D face spoofing database.
\subsection{Performance evaluation metrics}
Based on the ISO/IEC metrics, for the evaluation of the vulnerability of FRSs, the Impostor Attack Presentation Match Rate (IAPMR) metric was used to report the results, which can be considered as an indication of the attack success chances if the FRS is evaluated regarding its PAD capabilities~\cite{scherhag2017v}. It is defined as the proportion of impostor attack presentations using the same Presentation Attack Instrument (PAI) species in which the target reference is matched in a full-system evaluation of a verification system. 

For the detection performance evaluation, we reported the results using the Attack Presentation Classification Error Rate (APCER), the Bona Fide Presentation Classification Error Rate (BPCER), and the Average Classification Error Rate (ACER). They are calculated as follows:

\begin{small}
\begin{equation}
\setlength{\abovedisplayskip}{5pt}
\setlength{\belowdisplayskip}{3pt}
APCER=\frac{1}{N_{a}}\sum_{i=1}^{N_{a}}(1-Res_{i})
\end{equation}
\begin{equation}
\setlength{\abovedisplayskip}{3pt}
\setlength{\belowdisplayskip}{3pt}
BPCER=\frac{\sum_{i=1}^{N_{r}}Res_{i}}{N_{r}}
\end{equation}
\begin{equation}
\setlength{\abovedisplayskip}{3pt}
\setlength{\belowdisplayskip}{3pt}
ACER=\frac{APCER+BPCER}{2}
\end{equation}
\end{small}
where ${N_{a}}$ is the total number of attack presentations, and ${N_{r}}$ is the number of real samples. ${Res_{i}}$ equals to 1 if the $i^{th}$ presentation is classified as an attack and 0 if classified as real. Lower values of these metrics indicate better performance of the PAD algorithms. 

\subsection{Vulnerabilities of face recognition systems}
Three FARs were considered to show the vulnerability towards detecting fake faces using the proposed WFFD database, so that the attack abilities of the super-realistic database can be demonstrated. They are two publicly available FRSs: OpenFace~\cite{Openface} and Face++~\cite{Facepp}, and a commercial system Neurotechnology VeriLook SDK~\cite{Verilook}. Using the thresholds recommended by these FRSs, we calculated the IAPMR values on three protocols of the WFFD database, as presented in Table 3.

\begin{table}[h]
\renewcommand{\arraystretch}{1.10}
\setlength{\tabcolsep}{7pt}
\centering
\begin{threeparttable}   
  \caption{IAPMR of three Face Recognition Systems}
    \begin{tabular}{|l|c|c|c|}
    \hline
    \textbf{Protocol}  & \textbf{Openface} & \textbf{Face++} &\textbf{VeriLook} \\
    \hline
    Threshold & 0.99\tnote{*} &1e-5\tnote{\dag} &36\tnote{**}\\
    \hline
    Protocol I& 93.29\%& 92.60\%  & 76.14\%  \\ 
   \hline
    Protocol II& 96.73\% & 96.22\% & 88.12\%  \\ 
   \hline
    Protocol III & 95.25\%  &94.72\% & 81.75\% \\
    \hline
\end{tabular}
\begin{tablenotes}
\scriptsize
\item[*] Using a squared L2 distance threshold; \dag Using the confidence threshold at the 0.001\% error rate; ** Using the matching score when FAR=0.1\%.
\end{tablenotes}
\end{threeparttable}
\end{table}
Table 3 shows that over 92\% of the images in the three protocols of the WFFD were successfully compared using the Openface and Face+, which means the high attack success chances of the proposed WFFD database on these two face recognition systems. However, lower values of the IAPMR can be seen when the VeriLook SDK was employed on the three protocols. This is attributed to the fact that some faces with special poses or low qualities cannot be identified by the VeriLook SDK, therefore, leading to less successful matches. 

In addition, by comparing the results for Protocol I and Protocol II, we can observe that higher IAPMR values were achieved for images in Protocol II, where the fake faces and real faces were recorded in the same scenarios with the same cameras. This leads to the higher attack abilities of images in Protocol II.

\subsection{Detection performance of face PAD algorithms}
Several face PAD methods were evaluated on the WFFD database to show how they can work for the proposed super-realistic 3D presentation attacks. These PAD methods were based on different features, including the multi-scale LBP~\cite{3DMAD2013spoofing}, the color LBP~\cite{A12boul2015face}, the Haralick features~\cite{agarwal2016face}, the reflectance properties~\cite{MORPHO4kose2013reflectance}, the multi-level Local Phase Quantization (LPQ)~\cite{A11benl2015face}, deep features based on the ResNet-50 model~\cite{D3tu2017ultra} and VGG-16 model~\cite{lucena2017transfer}. They all achieved high detection performance against 2D spoofing attacks or 3D mask attacks. 

In this experiment, for each image in the WFFD database, the two face regions were first detected, cropped and normalized into 64*64 pixel images. Based on different PAD methods, features were extracted from the face images, and then fed into a Softmax classifier with a cross-entropy loss function.

The detection results on Protocol I of the proposed WFFD are shown in Table 4. We can see that existing seven face PAD methods achieved high detection error rates for this protocol, ranging from 17\% to 46\%. We attribute the poor performances to the high diversity and super-realistic attacks in the WFFD database, therefore, making it difficult to detect real faces from wax figure faces recorded in different scenarios. Overall, due to the highly discriminative features learned by the ResNet-50 models, the method~\cite{D3tu2017ultra} achieved the best results, with the ACER of 19.27\%. The multi-level LPQ~\cite{A11benl2015face} based method also performed relatively well, with the ACER of 25.12\%.
\begin{table}[h]
\renewcommand{\arraystretch}{1.10}
\setlength{\tabcolsep}{3pt}
\centering
 \caption{Detection error rates (\%) of several PAD methods on the Protocol I of the WFFD}
    \begin{tabular}{|l|c|c|c|c|}
    \hline
    \textbf{Methods}  & \textbf{EER} & \textbf{APCER} &\textbf{BPCER} & \textbf{ACER} \\
    \hline
    Multi-scale LBP~\cite{3DMAD2013spoofing}&33.17  & 31.22 &  31.22  & 31.22\\ 
   \hline
    Color LBP~\cite{A12boul2015face} &  33.17 & 30.24 & 36.10  &33.17 \\ 
   \hline
    Haralick features~\cite{agarwal2016face} & 32.19 & 25.85 &  37.07 &  31.46\\ 
   \hline
    Reflectance~\cite{MORPHO4kose2013reflectance} &41.95  & 40.00  & 52.19 &  46.10\\ 
   \hline
    Multi-level LPQ~\cite{A11benl2015face}& 24.88 & 24.88 & 25.36 & 25.12 \\ 
   \hline
    ResNet-50 based~\cite{D3tu2017ultra} & \textbf{17.07} & \textbf{20.49} & \textbf{18.04} &  \textbf{19.27} \\ 
   \hline
    VGG-16 based~\cite{lucena2017transfer}&45.85 &  50.73  & 41.95  & 46.34\\
       \hline
\end{tabular}
\begin{tablenotes}
\scriptsize
\item[*]The best results are indicated in bold.
\end{tablenotes}
\end{table}

Table 5 presents the detection results using Protocol II on the WFFD database. Compared with the results in  Table 4, we can see that most error rates for this protocol are higher than the values for Protocol I. Such results are reasonable since recording the real faces and wax figure faces in the same scenarios with the same cameras results in less differences of the faces. Therefore, it is more difficult to detect the presentation attacks in this protocol, also meaning the higher attack abilities, which reached a similar conclusion with the results in Table 3.

Similar to the results in Table 4, the ResNet-50 based PAD method performed the best, achieving 21.83\% ACER, while the reflectance properties based ~\cite{MORPHO4kose2013reflectance} and VGG-16 based~\cite{lucena2017transfer} method have no advantages in distinguishing real faces from wax figure faces.

\begin{table}[h]
\renewcommand{\arraystretch}{1.10}
\setlength{\tabcolsep}{3pt}
\centering
  \caption{Detection error rates (\%) of several PAD methods on the Protocol II of the WFFD}
    \begin{tabular}{|l|c|c|c|c|}
    \hline
    \textbf{Methods}  & \textbf{EER} & \textbf{APCER} &\textbf{BPCER} & \textbf{ACER} \\
    \hline
    Multi-scale LBP~\cite{3DMAD2013spoofing}&36.62  & 37.32  & 33.45  & 35.39 \\ 
   \hline
    Color LBP~\cite{A12boul2015face} & 37.32  & 36.62  & 41.90  & 39.26  \\ 
   \hline
    Haralick features~\cite{agarwal2016face} &38.38 & 41.55 & 24.65 & 33.10  \\ 
   \hline
    Reflectance ~\cite{MORPHO4kose2013reflectance} &44.37 &  50.70  & 44.37 &  47.53 \\ 
   \hline
    Multi-level LPQ~\cite{A11benl2015face}&33.10  & 35.56  & 22.89  & 29.22\\ 
   \hline
    ResNet-50 based~\cite{D3tu2017ultra} &\textbf{20.42} & \textbf{19.37}  & \textbf{24.29}  & \textbf{21.83}  \\ 
   \hline
    VGG-16 based~\cite{lucena2017transfer}& 48.94 & 40.14 & 52.82 &  46.48  \\
       \hline
\end{tabular}
\begin{tablenotes}
\scriptsize
\item[*] The best results are indicated in bold.
\end{tablenotes}
\end{table}

The overall results of the seven evaluated PAD methods on the WFFD database are shown in Table 6. The ACER values ranged from 20.04\% to 47.24\%, showing the poor detection performance of these methods against the proposed wax figure face presentation attacks, and therefore, the strong attack abilities of the WFFD database in face recognition.
\begin{table}[h]
\renewcommand{\arraystretch}{1.10}
\setlength{\tabcolsep}{3pt}
\centering
  \caption{Detection error rates (\%) of several PAD methods on the Protocol III of the WFFD}
    \begin{tabular}{|l|c|c|c|c|}
    \hline
    \textbf{Methods}  & \textbf{EER} & \textbf{APCER} &\textbf{BPCER} & \textbf{ACER} \\
    \hline
    Multi-scale LBP~\cite{3DMAD2013spoofing}& 34.56 &  33.33  & 32.92  & 33.13 \\ 
   \hline
    Color LBP~\cite{A12boul2015face} & 36.81  & 35.38  & 35.79 & 35.58  \\ 
   \hline
    Haralick features~\cite{agarwal2016face} &  36.81  & 36.40  & 32.92 &  34.66 \\ 
   \hline
    Reflectance~\cite{MORPHO4kose2013reflectance} &  44.78&   46.01 &  46.22  & 46.11\\ 
   \hline
    Multi-level LPQ~\cite{A11benl2015face}& 28.63  & 29.45 &  25.15  & 27.30  \\ 
   \hline
    ResNet-50 based~\cite{D3tu2017ultra} & \textbf{18.81}  & \textbf{19.43}  & \textbf{20.65} & \textbf{20.04}  \\ 
   \hline
    VGG-16 based~\cite{lucena2017transfer}&48.67 &  45.19 &  49.28  & 47.24   \\
       \hline
\end{tabular}
\begin{tablenotes}
\scriptsize
\item[*] The best results are indicated in bold.
\end{tablenotes}
\end{table}

\section{Conclusion}

To address the limitations in existing 3D face presentation attack databases, we have proposed a new database, WFFD, composed of wax figure images with high diversity and large data size as super-realistic face presentation attacks. The database will be made publicly available in order to help the development and fair evaluation of different PAD algorithms. Extensive experiments have demonstrated the vulnerability of popular face recognition systems to these attacks, and the performance degradation of several existing PAD methods in detecting real faces from wax figure faces, showing the challenges when wax figure face are used for 3D attacks.

Some motion based methods, such as head movement and blink detection based may seem quite effective in detecting wax figure fake faces if the faces are recorded in videos. However, nowadays the high-tech wax figure technologies have realized the intelligent wax figures, which can not only move but also sense people and change its behaviour based on its surroundings. Therefore, it is demanding to investigate more discriminative and powerful methods to detect these new challenging 3D face presentation attacks in the future. 

\section*{Acknowledgment}
This work was partly supported by an NSF-CITeR project and a WV-HEPC grant, and Applied Basic Research Program of Wuhan (No.2017010201010114).

{\small
\bibliographystyle{ieee}
\bibliography{submission_example}
}

\end{document}